\documentclass[sigconf]{acmart}
\usepackage{multirow} 
\usepackage{subcaption}
\usepackage{makecell}

\AtBeginDocument{%
  \providecommand\BibTeX{{%
    Bib\TeX}}}

\setcopyright{acmlicensed}
\copyrightyear{2026}
\acmYear{2026}
\acmDOI{XXXXXXX.XXXXXXX}
\acmConference[Conference acronym 'XX]{Make sure to enter the correct
  conference title from your rights confirmation email}{June 03--05,
  2018}{Woodstock, NY}
\acmISBN{978-1-4503-XXXX-X/2018/06}

\begin{document}

\title{SPECTRA: Spectral Domain-Aware Graph Generation for Imbalanced Molecular Property Regression}

\author{Brenda Nogueira}
\email{bcruznog@nd.edu}
\orcid{0000-0003-1936-9863}
\affiliation{%
  \institution{University of Notre Dame,
  Dept. of Computer Science and Engineering}
  \city{Notre Dame}
  \state{Indiana}
  \country{USA}
}

\author{Gisela A. González-Montiel}
\email{ggonzal6@nd.edu}
\orcid{0000-0003-0665-5326}
\affiliation{%
  \institution{University of Notre Dame, Dept. of Chemistry and Biochemistry}
  \city{Notre Dame}
  \state{Indiana}
  \country{USA}
}

\author{Meng Jiang}
\email{mjiang2@nd.edu}
\orcid{0000-0002-3009-519X}
\affiliation{%
\institution{University of Notre Dame,
  Dept. of Computer Science and Engineering}
  \city{Notre Dame}
  \state{Indiana}
  \country{USA}
}

\author{Nitesh V. Chawla}
\email{nchawla@nd.edu}
\orcid{0000-0003-3932-5956}
\affiliation{%
\institution{University of Notre Dame,
  Dept. of Computer Science and Engineering}
  \city{Notre Dame}
  \state{Indiana}
  \country{USA}
}

\author{Nuno Moniz}
\email{nunomoniz@nd.edu}
\orcid{0000-0003-4322-1076}
\affiliation{%
 \institution{University of Notre Dame,
  Lucy Family Institute for Data \& Society}
  \city{Notre Dame}
  \state{Indiana}
  \country{USA}
}

\renewcommand{\shortauthors}{Nogueira et al.}

\begin{abstract}

  Molecular property regression struggles with cases in chemically-relevant target ranges that are underrepresented in datasets. Standard average error minimization approaches underperform in these highly relevant cases, and oversampling approaches lead to meaningless molecular representations. In this paper, we propose SPECTRA, a spectral, domain-aware graph generation method designed to improve the prediction of underrepresented but relevant molecular property values. 
  It combines a rarity-aware budgeting scheme to focus generation where data are scarce, target-neighbors graph alignment to establish structural correspondence, and interpolation of Laplacian spectra, node features, and targets. Coupled with spectral GNN using edge-aware Chebyshev convolutions, SPECTRA shows its effectiveness in property prediction benchmarks with competitive performance over leading state-of-the-art methods in relevant target ranges, while requiring \textasciitilde 4x less computational time.

\end{abstract}




\maketitle

\section{Introduction}
Graph-structured data plays a central role in many scientific domains, including drug discovery, materials science, and genomics. These fields produce large volumes of complex, structured information that can be naturally represented as graphs, where nodes correspond to entities (e.g., atoms, molecules, genes) and edges capture their relationships (e.g., chemical bonds, interactions, and regulatory relationships). Graph Neural Networks (GNNs) have transformed the modeling of such data by operating directly on graph structures, enabling state-of-the-art predictions of molecular properties, material characteristics, and biological interactions. In drug discovery, for example, GNNs have been applied to property prediction \citep{xiong2019pushing}, molecular design \citep{jin2018junction}, and drug–target interaction prediction \citep{lim2019predicting}, with increasing adoption by the pharmaceutical industry to accelerate a development pipeline \citep{vamathevan2019applications}. Similar advances have been reported in materials science, where GNNs help identify compounds with desirable structural and functional properties~\citep{karamad2020orbital},  such as crystalline stability, band gaps, and mechanical strength.  
  
Despite this progress, a fundamental challenge remains unsolved: \emph{imbalanced regression} on graphs. While imbalanced classification has received significant attention in graph learning, the regression setting has been comparatively neglected \citep{almeida2024overcoming, xia2024novel,ribeiro2020imbalanced, liu2023semi}. Yet, many scientific problems involve continuous targets where the most valuable information comes from rare samples. Standard GNNs and other machine learning methods typically optimize for average performance across the full label distribution, which leads to poor accuracy in these rare but scientifically important regions. Common data augmentation methods with graph contrastive learning (e.g., edge perturbation, node dropping) are inadequate since they destroy biochemical structures.~\citep{zhang2025decoydb,wang2022molecular, magar2022auglichem}.  Another challenge lies in \emph{embedding-based augmentation methods}, which often generate synthetic molecules in latent spaces that lack interpretability and offer no guarantees of structural validity. As a result, the augmented samples will unlikely correspond to chemically realistic molecules. 

To address these challenges, we propose \textbf{SPECTRA} -- \emph{Spectral Domain-Aware Graph Generation for Imbalanced Molecular Property Regression}. SPECTRA introduces a novel approach to oversampling that operates directly in the spectral domain of graphs. Specifically, it leverages the eigenspace of the graph Laplacian to interpolate both Laplacian spectra and node features of matched graphs in a shared spectral basis. Unlike black-box embedding methods, this process produces molecular graphs that are physically coherent, chemically plausible, and explicitly tailored to underrepresented regions of the target distribution (see Figure~\ref{fig:dist}), 
and achieving considerably lower computational cost w.r.t. state-of-the-art performing methods. Our contributions are as follows:

\begin{figure}
    \centering
    \includegraphics[width=\linewidth]{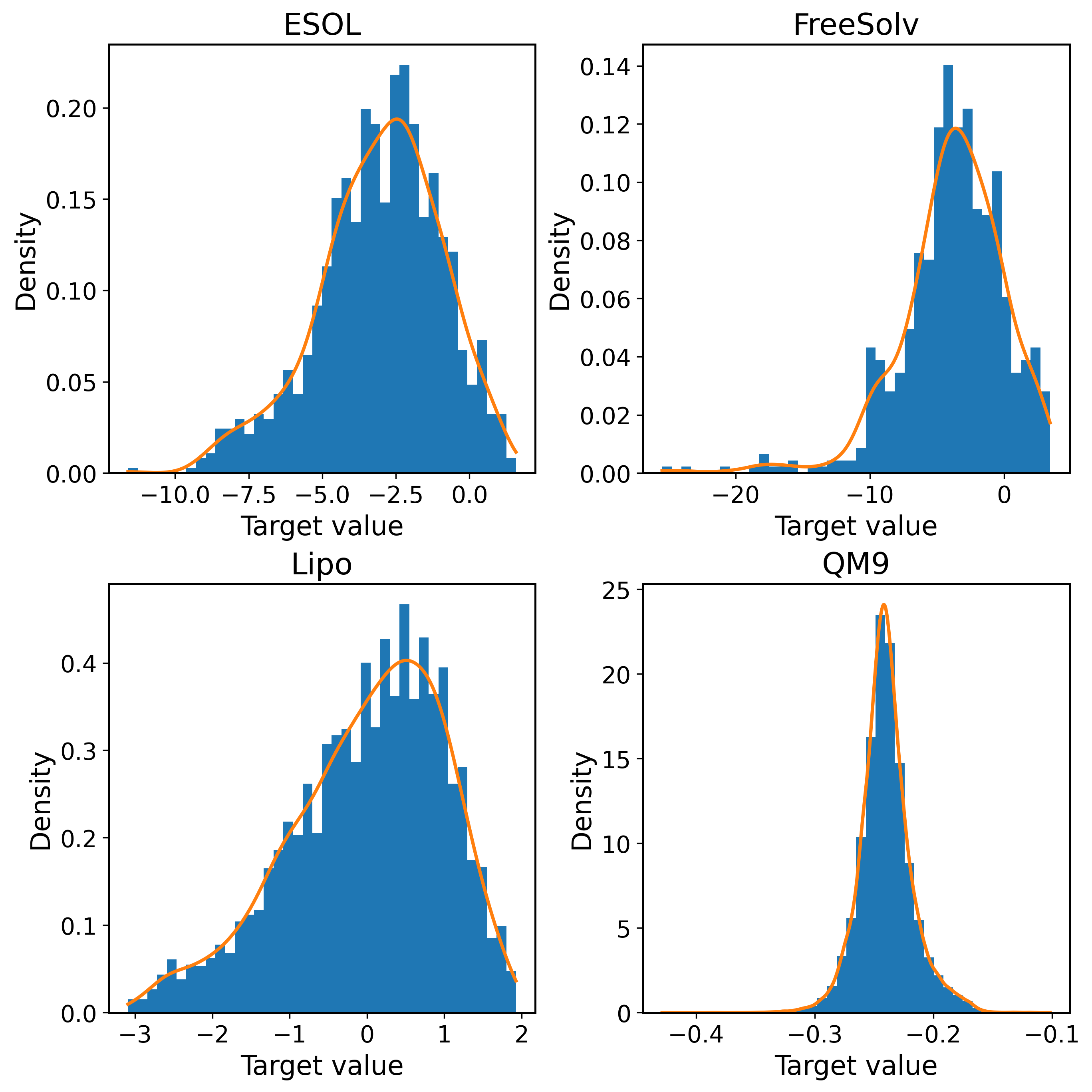}
    \caption{Distribution of target property values across three molecular datasets (ESOL, FreeSolv, Lipo, and QM9 (homo property). Each subplot shows a normalized histogram of the experimental values with a Gaussian kernel density estimate (KDE) overlaid using Scott’s rule-of-thumb bandwidth. These plots highlight the skewness and spread of target distributions, which can influence model training and performance.}
    \label{fig:dist}
\end{figure}  

\begin{itemize}
    \item \textbf{Novel methodology.} We introduce a spectral-based generative method that generates samples in low-density regions of the label space while preserving topological fidelity, overcoming the limitations of existing oversampling techniques in regression.  
    \item \textbf{Improved predictive performance.} Across benchmark molecular property datasets, SPECTRA achieves low error on rare compounds without degrading performance on common cases.  
    \item \textbf{Interpretability and efficiency.} The synthetic graphs generated by SPECTRA are realistic and chemically meaningful, enabling direct inspection of generated molecules while maintaining a lower computational footprint compared to competing approaches.  
\end{itemize}  

All materials required to replicate the results presented in this paper are available in~\url{https://github.com/brendacnogueira/SPECTRA}.

\begin{figure*}
    \centering
    \includegraphics[width=0.9\linewidth]{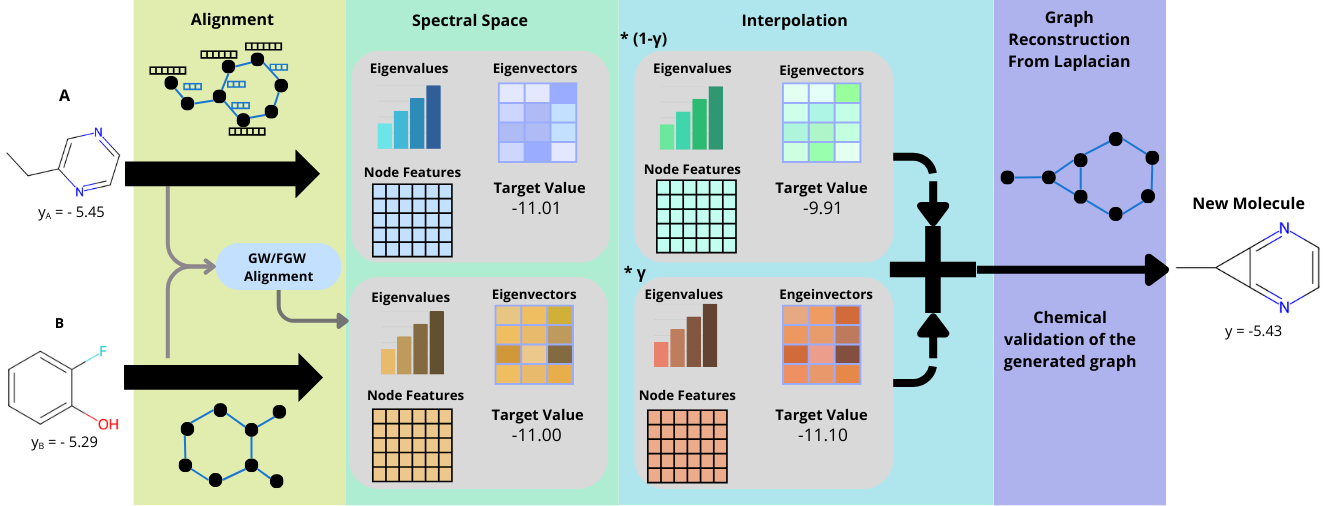}
    \caption{SPECTRA Pipeline. Molecular graphs are first aligned using Gromov–Wasserstein matching, which maps graph B into the structural space of graph A. Their Laplacians are then decomposed into eigenvalues and eigenvectors to obtain a spectral representation. Interpolation is performed in this spectral space-along with node features and target values—by applying a linear combination to each component, producing an interpolated spectral representation. Finally, the resulting Laplacian and node embeddings are used to reconstruct the interpolated molecular graph, which follows basic principles of chemistry.}
    \label{fig:pipeline}
\end{figure*}

\section{Related Work}

The challenge of imbalanced distributions in graph learning tasks has received increasing attention, particularly in scientific domains where rare values are critical. Recent research by \citet{almeida2024overcoming} demonstrates that imbalanced learning in drug discovery datasets can be tackled with techniques such as oversampling and loss function manipulation when using Graph Neural Networks (GNNs). Despite these advances, most approaches operate directly in graph space rather than the spectral domain, limiting their ability to maintain global structural constraints. \citet{bo2023survey} published a comprehensive survey on spectral GNNs, highlighting their unique ability to capture global information and provide better expressiveness than spatial approaches. \citet{wang2022powerful} further analyzed the theoretical expressive power of spectral GNNs, proving that they can produce arbitrary graph signals under specific conditions. However, these methods focus on balanced and classification datasets, illustrating the novelty and significance of SPECTRA.

\subsection{Imbalanced Learning}

Class imbalance has traditionally been addressed through resampling strategies, such as under-sampling majority classes or over-sampling minority classes. SMOTE~\citep{chawla2002smote}, for instance, generates synthetic minority samples by interpolating labeled data. Alternative approaches include cost-sensitive learning~\citep{cui2019class,lin2017focal}, which increases the loss weight of minority classes, and posterior recalibration~\citep{cao2019learning,menon2020long,tian2020posterior}, giving larger margins for minority predictions. In the context of a graph, a lot of approaches have been developed for imbalanced classification~\citep{maGraphClassImbSurvey}.

Imbalanced regression introduces additional challenges because labels are continuous rather than categorical~\citep{ribeiro2020imbalanced}. Several methods from classification have been adapted to this setting. For example, SMOGN~\cite{branco2017smogn} extends SMOTE to regression, while BMSE~\cite{ren2022balanced} adapts logit re-calibration for numerical targets. LDS~\cite{yang2021delving} smooths the label distribution using kernel density estimation, and RankSim~\cite{gong2022ranksim} regularizes the latent space by aligning distances in label and feature space. Other approaches include SERA~\cite{ribeiro_imbalanced_2020}, which proposes a relevance-aware evaluation metric; and SIRN~\cite{zong2024boosting}, which combines deviation modeling with adaptive pseudo-label selection. In the graph domain, SGIR~\cite{liu2023semi} leverages unlabeled graphs to enrich underrepresented label ranges.  While these methods improve performance in underrepresented regions, they often reduce accuracy in well-represented areas, especially under limited supervision or when relying heavily on pseudo-labeling.

\subsection{Spectral Graph Methods}

Spectral graph theory has applications spanning dimensionality reduction, clustering, and graph signal processing. Recent work in spectral methods includes Specformer~\citep{bo2023specformerspectralgraphneural}, combining spectral GNNs with transformer architectures to create learnable set-to-set spectral filters, or the work by~\citet{ding2025large} to enhance the scalability of spectral GNNs without decoupling the network architecture, addressing a key limitation in previous approaches.~\citet{yang2024spectral} present a spectral-aware augmentation method that selectively perturbs eigenpairs to preserve task-relevant frequency bands in graph contrastive learning. These advanced spectral methods demonstrate improved performance on various graph learning tasks, but do not specifically target the regression setting or leverage the spectral domain for learning in imbalanced scenarios. 
\citet{alharbi2024spectral} applied spectral graph convolutional neural networks to Alzheimer's disease diagnosis, showing how these representations can capture complex relationships in biomedical data, but did not address the challenge of imbalanced target distributions.



\subsection{Graph Sampling and Synthesis in Scientific Domains}

Due to domain-specific constraints and validity requirements, scientific applications pose unique challenges for graph-based methods. \citet{yao2024knowledge} provided a comprehensive bibliometric analysis of GNN applications in drug discovery, showing significant growth in this area and highlighting the need for methods to handle the inherent data imbalances in these domains. Similarly, \citet{fan2024reducing} addressed the challenge of overconfident errors in molecular property classification, demonstrating the importance of uncertainty quantification in imbalanced datasets. These approaches focus primarily on classification rather than regression tasks. On regression tasks, a review on GNNs for predicting synergistic drug combinations~\citep{besharatifard2024review} noted that graph-based models often suffer from imbalanced data distributions, affecting their performance. They emphasized the need for methods to handle such imbalances to improve predictive accuracy effectively.

\subsection{Molecular Generation}

Molecular generation has become a central task in drug discovery, aiming to explore chemical space efficiently while ensuring chemical validity and optimizing for desired properties. Early approaches combined variational autoencoders (VAEs), recurrent neural networks (RNNs), and adversarial models to generate novel chemical structures from latent spaces, as in LatentGAN~\citep{prykhodko2019novo}, which integrated autoencoding with generative adversarial training for de novo molecular design. More recent methods leverage reinforcement learning to incorporate chemical constraints and multi-objective optimization. For example, DeepGraphMolGen~\citep{khemchandani2020deepgraphmolgen} employs Graph Convolutional Policy Networks to generate molecules while simultaneously optimizing for drug-likeness and synthetic accessibility, whereas MORLD~\citep{jeon2020autonomous} integrates reinforcement learning with docking simulations to propose inhibitors directly guided by protein structures. Conditional generative frameworks, such as MGCVAE~\citep{lee2022mgcvae}, enable property-conditioned molecular graph generation, allowing for inverse design tasks like optimizing logP or molar refractivity. Beyond purely graph-based approaches, protein-informed generation methods such as DeepTarget~\citep{chen2023deep} directly construct candidate molecules from amino acid sequences of target proteins, bridging structural biology with generative chemistry. Diffusion models have also been widely adopted in molecular generation~\cite{zhang2023survey}, for example, in DiGress~\cite{vignac2022digress}, discrete diffusion process that progressively edits graphs with noise, through the process of adding or removing edges and changing the categories. Despite these advances, most existing models focus on validity, novelty, and property optimization, without explicitly addressing the imbalance of molecular property distributions.

\paragraph{\textbf{Novelty of SPECTRA}} Our method introduces a spectral-based generative strategy that explicitly targets underrepresented regions of the label space while preserving global graph structure and chemical validity. Unlike many existing approaches that rely on pseudo-labeling or sacrifice accuracy in well-represented regions, SPECTRA generates new, chemically coherent samples where data are sparse, mitigating imbalance without degrading overall performance. By combining spectral alignment with rare-target–aware sampling and validity reconstruction, it enables computationally-efficient interpretable molecule generation and improves prediction in rare but scientifically critical regimes.

\section{Method}\label{sec:methods}

We propose a spectral, domain-aware generative and learning pipeline for molecular property prediction that (i) constructs multi-attribute Laplacian representation from molecular graphs; (ii) aligns laplacians and nodes representations of different graphs using (Fused) Gromov–Wasserstein (FGW)~\citep{memoli2011gromov,vayer2020fused} couplings; (iii) interpolates eigenvalues and eigenvectors, along to node features in a stable orthonormal basis; and (iv) trains a spectral GNN with edge-aware Chebyshev convolutions~\citep{defferrard2016convolutional} on original and augmented samples. Figure~\ref{fig:pipeline} summarizes the workflow.

\subsection{From SMILES to Multi-Attribute Graphs}
Given a SMILES string $s$, we construct a graph $G=(V,\mathbf{X},\mathbf{E},y)$ with RDKit\footnote{RDKit: \url{https://www.rdkit.org}}, where $|V|$ is a set of nodes (atoms), $\mathbf{X}\in\mathbb{R}^{n\times d}$ atom-feature matrix (using OGB utilities), with $n$ being number of nodes and $d$ node features dimensionality, and each undirected edge $(u,v)\in E$, for $u,v \in V $, has a 3D attribute vector (bond type, stereo, conjugation). We treat the three edge channels as separate weighted adjacencies $\{\mathbf{W}^{(f)}\}_{f=1}^F$ ($F$ is the number of edge-attribute channels) and we compute a (unnormalized) Laplacian per channel:
\[
\mathbf{L}^{(f)}=\mathbf{D}^{(f)}-\mathbf{W}^{(f)},\quad
\mathbf{D}^{(f)}=\operatorname{diag}\!\big(\mathbf{W}^{(f)}\mathbf{1}\big),
\]
\noindent where $\mathbf{1}$ is the vector of ones used for degree computation.
To prevent disconnected subgraphs within individual bond-type channels, we add a small constant offset to each corresponding adjacency matrix, ensuring that all nodes maintain minimal connectivity. This adjustment stabilizes the Laplacian spectra across channels and avoids degenerate eigenmodes during spectral interpolation while preserving the relative structural topology. After generating the interpolated Laplacian, the offset is subtracted to restore the original structural scale before reconstruction.

\subsection{Geometry-Aware Graph Alignment (FGW)}\label{sec:methods-align}

Network alignment (or graph matching) is a widely studied problem across scientific domains. In its basic form, it aims to identify a node correspondence between two graphs that best preserves their structural relationships~\cite{lazaro2025probabilistic}. In molecular applications, this alignment enables the comparison of structural similarity across different compounds~\cite{emmert2016fifty}.

To establish node correspondence between two molecules $A$ and $B$, we solve a Gromov–Wasserstein (GW) or Fused GW (FGW) optimal transport problem on their zero-padded adjacency matrices $\tilde{\mathbf{A}},\tilde{\mathbf{B}}\in\mathbb{R}^{n\times n}$ (padding each graph to the larger node count).   We define probability distributions $p,q\in\Delta^n$ (set of valid probability distributions) over the nodes of $A$ and $B$, respectively.  
Each entry $p_i$ (or $q_j$) represents the relative “mass’’ assigned to node $i$ in $A$ (or node $j$ in $B$); in this work we use uniform weights so that $p_i = 1/|V_A|$ and $q_j = 1/|V_B|$.  
The transport plan $\mathbf{T}\in\mathbb{R}^{n\times n}$ then specifies how this probability mass is moved from each node of $A$ to each node of $B$, effectively giving a soft alignment between their nodes.  
When node attributes are available, we use FGW with a cost matrix $\mathbf{M}$ that measures feature dissimilarity; otherwise we use pure GW:

\[
\mathbf{T}^\star
=\arg\min_{\mathbf{T}\in\Pi(p,q)}
(1-\alpha)\,\mathcal{L}_{\mathrm{GW}}(\tilde{\mathbf{A}},\tilde{\mathbf{B}},\mathbf{T})
+\alpha\,\langle \mathbf{M},\mathbf{T}\rangle,
\]
\noindent
where $\Pi(p,q)$ is the set of couplings with marginals $p$ and $q$, $\mathcal{L}_{\mathrm{GW}}$ is the squared-loss GW discrepancy, and $\alpha\in[0,1]$ balances structural versus feature similarity.  
The resulting optimal coupling $\mathbf{T}^\star$ provides a soft node-to-node correspondence; we convert it into a hard one-to-one mapping using the Hungarian assignment on $-\mathbf{T}^\star$ and reorder $\tilde{\mathbf{B}}$ and its features accordingly.

\subsection{Spectral Interpolation} \label{sec:methods-interpolation}
Given the matched pair, $\boldsymbol{\Lambda}_A, \boldsymbol{\Lambda}_B$ being the eigenvalues of A and B, respectively, and $\mathbf{U}_A, \mathbf{U}_B$ being the eigenvectors of A and B, respectively, we diagonalize
\[
\mathbf{L}_A=\mathbf{U}_A\boldsymbol{\Lambda}_A\mathbf{U}_A^\top,\quad
\mathbf{L}_B=\mathbf{U}_B\boldsymbol{\Lambda}_B\mathbf{U}_B^\top,
\]
and align eigenvector signs and bases with an orthogonal Procrustes map
$\mathbf{R}^\star=\arg\min_{\mathbf{R}\in O(k)}\|\mathbf{U}_A^\top\mathbf{U}_B-\mathbf{R}\|_F$, given $
O(k) = \{\, R \in \mathbb{R}^{k \times k} \;:\; R^{\top} R = I \,\}
$ where $||\cdot||_F$ is the Frobenius norm, and $k$ is the number of eigenvalues/eigenvectors, yielding $\widetilde{\mathbf{U}}_B=\mathbf{U}_B\mathbf{R}^\star$. We then interpolate eigenvalues and bases with a mixing coefficient $\gamma\in(0,1)$:
\[
\boldsymbol{\Lambda}_\gamma=(1-\gamma)\boldsymbol{\Lambda}_A+\gamma\boldsymbol{\Lambda}_B,\quad
\widehat{\mathbf{U}}=(1-\alpha)\mathbf{U}_A+\gamma\,\widetilde{\mathbf{U}}_B,\quad
\mathbf{U}_\gamma=\operatorname{qr}(\widehat{\mathbf{U}}),
\]
where $qr(\cdot)$ factorization used for re-orthogonalization, and synthesize an intermediate Laplacian:
\[
\mathbf{L}_\gamma=\mathbf{U}_\gamma\,\boldsymbol{\Lambda}_\gamma\,\mathbf{U}_\gamma^\top.
\]
We repeat this per edge channel ($F{=}3$).

\paragraph{Node feature interpolation.}
In the matched node domain we perform linear interpolation in the original node space, where $\widetilde{\mathbf{X}}_B$ is $\mathbf{X}_B$ permuted by the GW/FGW correspondence:
\[
\mathbf{X}_\gamma=(1-\gamma)\,\mathbf{X}_A+\gamma\,\widetilde{\mathbf{X}}_B,
\]

\subsection{Graph Reconstruction from Spectra}
For each channel, we map $\mathbf{L}_\gamma^{(f)}$ back to a nonnegative adjacency:

\[
\mathbf{W}_\gamma^{(f)}=\max\!\big(0,\;-\mathbf{L}_\gamma^{(f)}+\operatorname{diag}(\mathbf{L}_\gamma^{(f)})\big),\quad
\operatorname{diag}(\mathbf{W}_\gamma^{(f)})=\mathbf{0}.
\]
We remove degrees and clip negatives. We then assemble multi-attribute edges by scanning $(u,v)$ with any positive channel weight and stacking per-channel features. The scalar label is interpolated as $y_\alpha=(1-\gamma)y_A+\gamma y_B$.

\subsection{Rarity-Aware Pair Selection and Augmentation Budget}
We compute a KDE over training labels to estimate density $\rho(y)$ and define rarity weights $w_i\propto 1/\rho(y_i)$ (normalized). Each training molecule $i$ receives an augmentation budget $\lfloor w_i\cdot N\cdot \texttt{perc}\rfloor$ (\texttt{perc}\,$\in[0,1]$ is a global rate). For molecule $i$, we sort neighbors by $|y_i-y_j|$ and generate pairs $(i,j)$ in that order, producing up to the allocated number of augmented graphs. 

For datasets with long tail label distributions, this process may cause $w_i \cdot N \cdot \texttt{perc}$ collapse to (near) zero, resulting in almost no augmentations for most samples. To address this, we use the number of augmented samples ($\lfloor N \cdot \texttt{perc} \rfloor$) and allocate counts using a multinomial draw with probabilities proportional to the inverse densities, ensuring a stable and non-degenerate augmentation budget.

\begin{table*}[!h]
    \centering
     \caption{Validity, uniqueness, novelty and internal diversity and Tanimoto similarity between original compounds of generated molecules for generated molecules across datasets.}\label{tab:molecules_genrated}
    \begin{tabular}{|l|l|c|c|c|c|c|}
    \hline
        \textbf{Dataset}  &  Model & $Validity\uparrow$ & $Unique\uparrow$ & $Novelty\uparrow$ & $Diversity\uparrow$ & $Similarity$\\ \hline
        \multirow{2}{*}{FreeSolv} & SPECTRA & 1.000 & 0.568 & 1.000 & 0.949 & 0.167 \\ 
                                  & VAE     & 1.000 & 0.404 & 0.900 & 0.879  & 0.671 \\ \hline
        \multirow{2}{*}{ESOL}  & SPECTRA & 1.000 & 0.661 & 0.949 & 0.929 & 0.365 \\ 
                               & VAE     & 1.000 & 0.781 & 0.880 & 0.881  & 0.487 \\ \hline
         
        \multirow{2}{*}{Lipo} & SPECTRA & 1.000 & 0.706 & 0.992 & 0.918 & 0.236  \\ 
                              & VAE  &   1.000 &  0.907  & 0.963  & 0.885  & 0.616  \\ \hline
        \multirow{2}{*}{QM9}  & SPECTRA & 1.000 & 0.928 & 0.819 &  0.930 &  0.680    \\
                              & VAE &  1.000  & 0.884 & 0.652  & 0.923 &  0.953 \\ 
        \hline
    \end{tabular}
\end{table*}
\subsection{Graph to Molecule}
\label{sec:graph-validity}

Interpolation of graphs gives no guarantee that it will yield valid molecules. 
Many generative frameworks produce very large sets of candidate molecules and subsequently filter or rank them based on synthesizability, stability, or task-specific criteria~\citep{gao2024generative,xu2022geodiff}. While not theoretically guaranteed, these assumptions have consistently been adopted in generative chemistry pipelines. 

To verify that the augmented graphs correspond to real chemical compounds, we convert each generated graph back into a SMILES string and validate its chemical consistency with RDKit. Each node’s feature vector is translated back into an atom description by mapping the numerical features to basic chemical properties: the atom type (e.g., carbon, oxygen), its charge, chirality, hybridization state, and whether it is aromatic. We do not consider radical molecules. Bonds are reconstructed from edge attributes, including type, stereochemistry, and conjugation, while avoiding duplicates to maintain a simple chemical graph.

The resulting editable molecule is then sanitized with RDKit to enforce valence rules, aromaticity perception, and proper connectivity.  Finally, the molecule is converted to a canonical SMILES string; graphs that cannot be sanitized are marked invalid. 


\subsection{Spectral GNN with Edge-Aware Chebyshev Convolutions}\label{sec:model}

We use a stack of spectral graph-convolutional block layers with Chebyshev filters (\texttt{ChebConv}~\cite{defferrard2016convolutional}), batch normalization, SiLU activation, and dropout. ChebConv provides localized polynomial approximations of Laplacian spectral filters, enabling efficient frequency-aware message passing that aligns naturally with SPECTRA’s spectral generation and structure-preserving interpolation objectives. Let $\mathbf{H}^{(0)}=\mathbf{X}$ denote the input node features, and $\mathbf{H}^{(\ell)}\in\mathbb{R}^{n\times d_\ell}$ the hidden representation at layer $\ell$. Given a graph operator $\mathbf{A}$ (constructed from the multi-channel Laplacians) and learnable edge-projection parameters $\mathbf{w}_e$, each block computes
\[
\mathbf{H}^{(\ell+1)}=
\operatorname{Drop}\!\left(
\operatorname{SiLU}\!\Big(
\operatorname{BN}\big(
\operatorname{ChebConv}_{K}(\mathbf{H}^{(\ell)},\,\mathbf{A},\,\mathbf{w}_e)
\big)
\Big)
\right),
\]
where $K$ is the Chebyshev polynomial order and $\operatorname{Drop}(\cdot)$ is applied with probability $p_{\mathrm{drop}}$. Each edge $(u,v)$ carries a 3-dimensional attribute vector $\mathbf{e}_{uv}\in\mathbb{R}^3$, which is projected through $\mathbf{w}_e$ and incorporated into the convolutional kernels.

\section{Experimental Evaluation}
We evaluate our methods across four benchmark datasets (FreeSolv, ESOL, Lipo and QM9), in detail in Appendix~\ref{app:dataset}. To assess the model’s ability to generate a diverse set of real molecules distinct from the training data (RQ1) we evaluated the generated drugs against Lipinski’s Rule of Five, a widely used guideline in drug discovery for assessing physicochemical properties essential for oral bioavailability~\cite{gimenez2010evaluation}, and assessed  other important properties:
quantitative estimate of drug-likeness (QED)~\citep{bickerton2012quantifying} and synthetic accessibility score (SAS)~\citep{ertl2009estimation}. 
as well as standard metrics like validity, uniqueness, and novelty. We also evaluate the impact of the different steps of our methods along with their different parameters (RQ2), and assess the predictive performance of our model against state-of-the-art methods (RQ3). 
We also assess the computational efficiency of our approach relative to existing methods (RQ4) and finally, we analyze its behavior across the entire target domain to understand improvements in low-density regions compared to high-density regions (RQ5). 
\begin{figure}[!h]
    \centering
    \includegraphics[width=\linewidth]{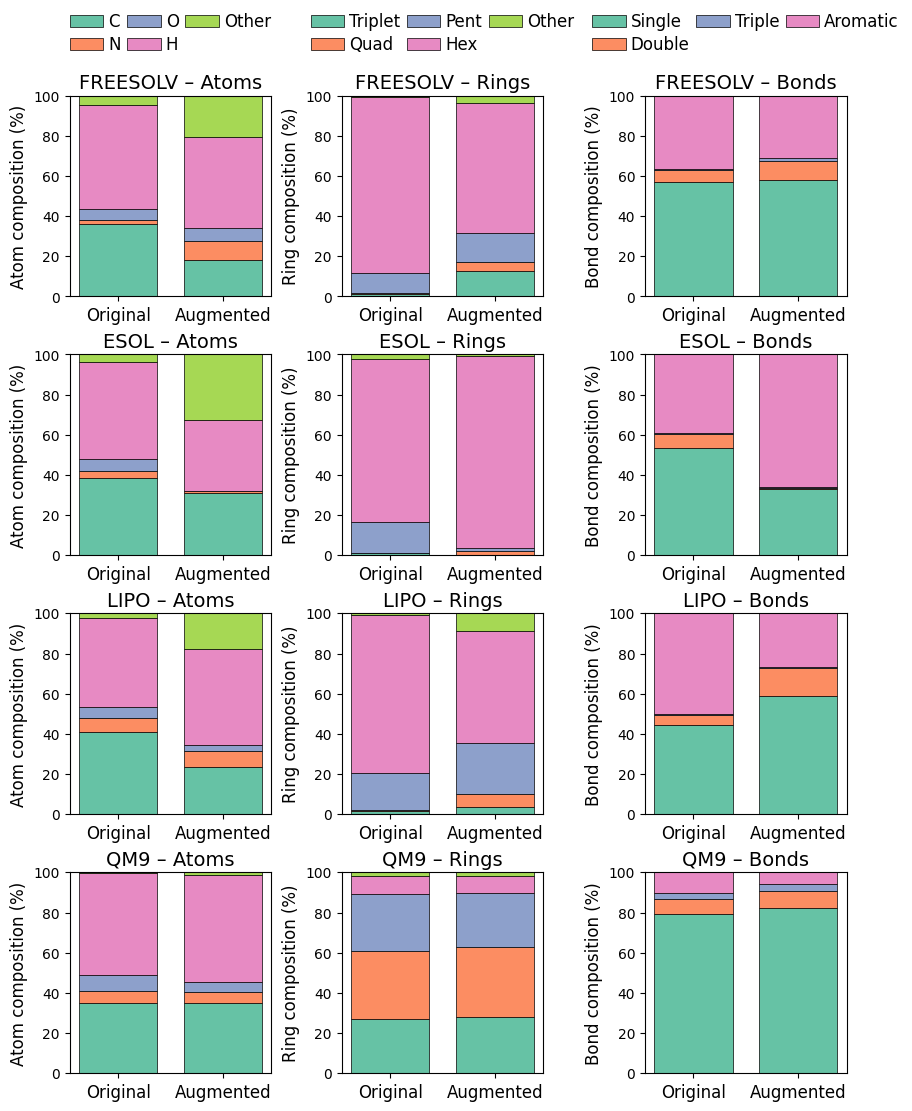}
    \caption{Comparison of the structural features between the original drugs from the FreeSolv, ESOL, Lipo  and QM9 datasets and the drugs generated by SPECTRA. The first column represents the percentage of different atoms (C=carbon, H=Hidrogen, N=nitrogen, O=oxygen, Other), the second column shows the proportion of different ring types (Tri, Quad, Pent, Hex, Other), while the third column reflects the diversities of bond types (Single, Double, Triple) in the molecules.}
    \label{fig:structural_feat}
\end{figure}

\subsection{Molecule Generation Quality (RQ1)}
To evaluate the quality of the generated molecules, we first assess their \emph{validity}, \emph{uniqueness}, \emph{novelty}, and \emph{Internal Diversity}. Validity is the fraction of generated molecules that follow basic principles of chemistry, i.e the ones that can be converted to a molecular structure. Uniqueness is the fraction of valid molecules that are non-duplicate, novelty is the fraction of unique molecules not present in the training set~\citep{flam2022language}, and internal diversity measures the structural variety \emph{within} the generated set, defined as one minus the mean pairwise Tanimoto similarity over Morgan fingerprints~\citep{rogers2010extended} of the unique generated molecules.

To evaluate whether the gains of our method stem from graph-level interpolation specifically, rather than from the augmentation strategy in general, we construct a VAE baseline~\citep{marchan2025vae} where new molecules are generated by perturbing each training molecule's posterior mean in a learned continuous latent space (Appendix~\ref{app:vae_motivation}).

Table~\ref{tab:molecules_genrated} shows that both methods achieve perfect validity across all datasets. It was expected of our method due to the validity-checking step. Spectra consistently outperforms the VAE in novelty across all benchmarks, with the largest gaps on QM9, where the VAE largely reconstructs known structures rather than generating new ones. Spectra also achieves higher internal diversity on all datasets, indicating that graph-level interpolation between structurally diverse parent pairs produces a more varied augmented set than neighborhood sampling in a learned latent space. The VAE achieves higher uniqueness only on ESOL and Lipo, but at the cost of lower novelty, suggesting its generated molecules, while distinct from each other, remain closer to the training set. Similarity with original molecules is lower than VAE in all datasets (< 0.7), suggesting structurally distinct molecules, expected in this imbalance regression setting, that reinforce the \textit{de novo} potential of our method.  On the contrary,  VAE has a higher similarity, suggesting closer generative molecules. Overall, \textbf{SPECTRA produces new, valid structures rather than replicating training molecules}, supporting the advantage of interpolating directly in graph space over traversing a continuous latent representation.

\begin{figure}[h]
\centering
        \begin{subfigure}[t]{0.49\linewidth}
            \centering
            \includegraphics[width=\linewidth]{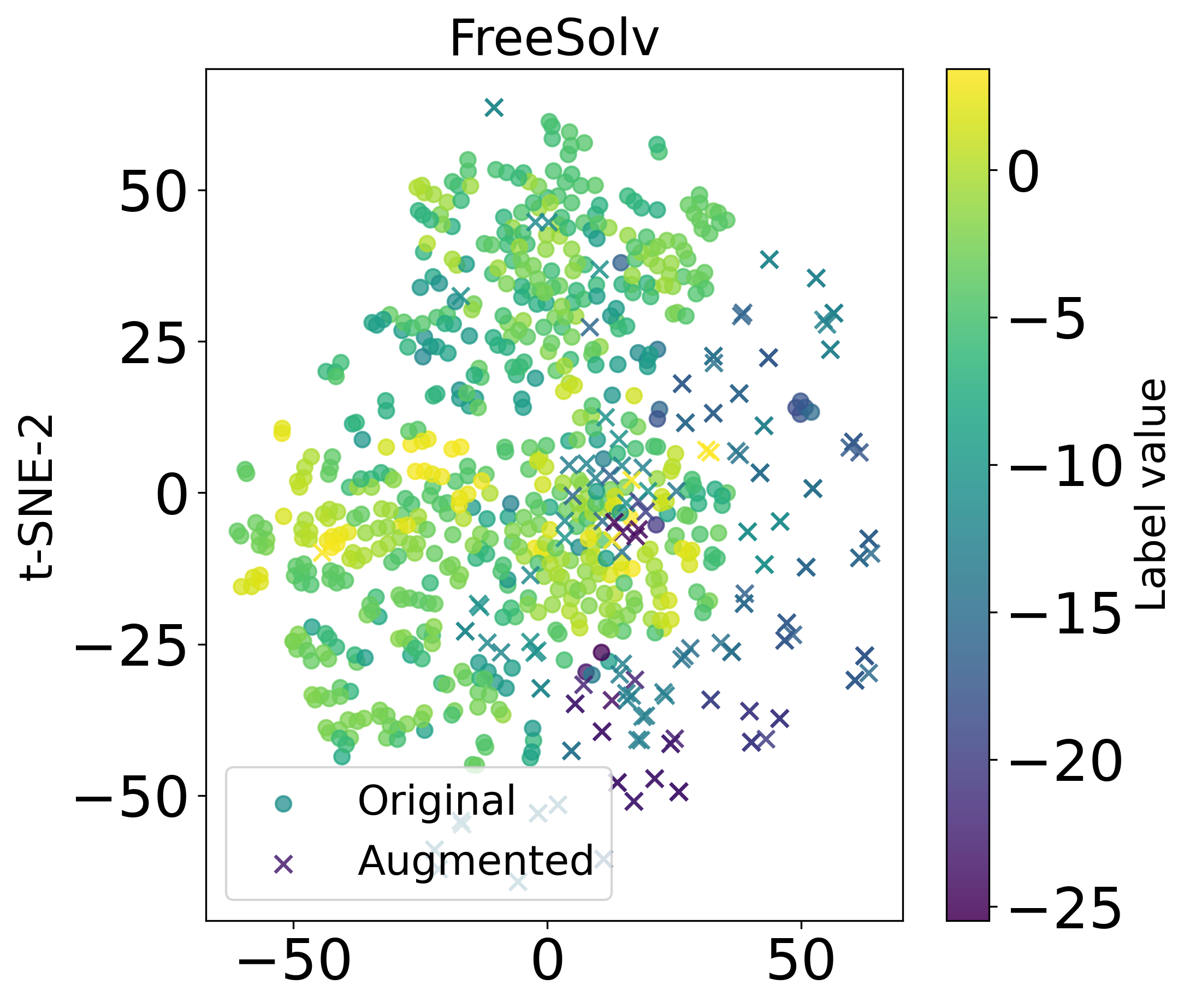}
        \end{subfigure}
        \begin{subfigure}[t]{0.49\linewidth}
            \centering
            \includegraphics[width=\linewidth]{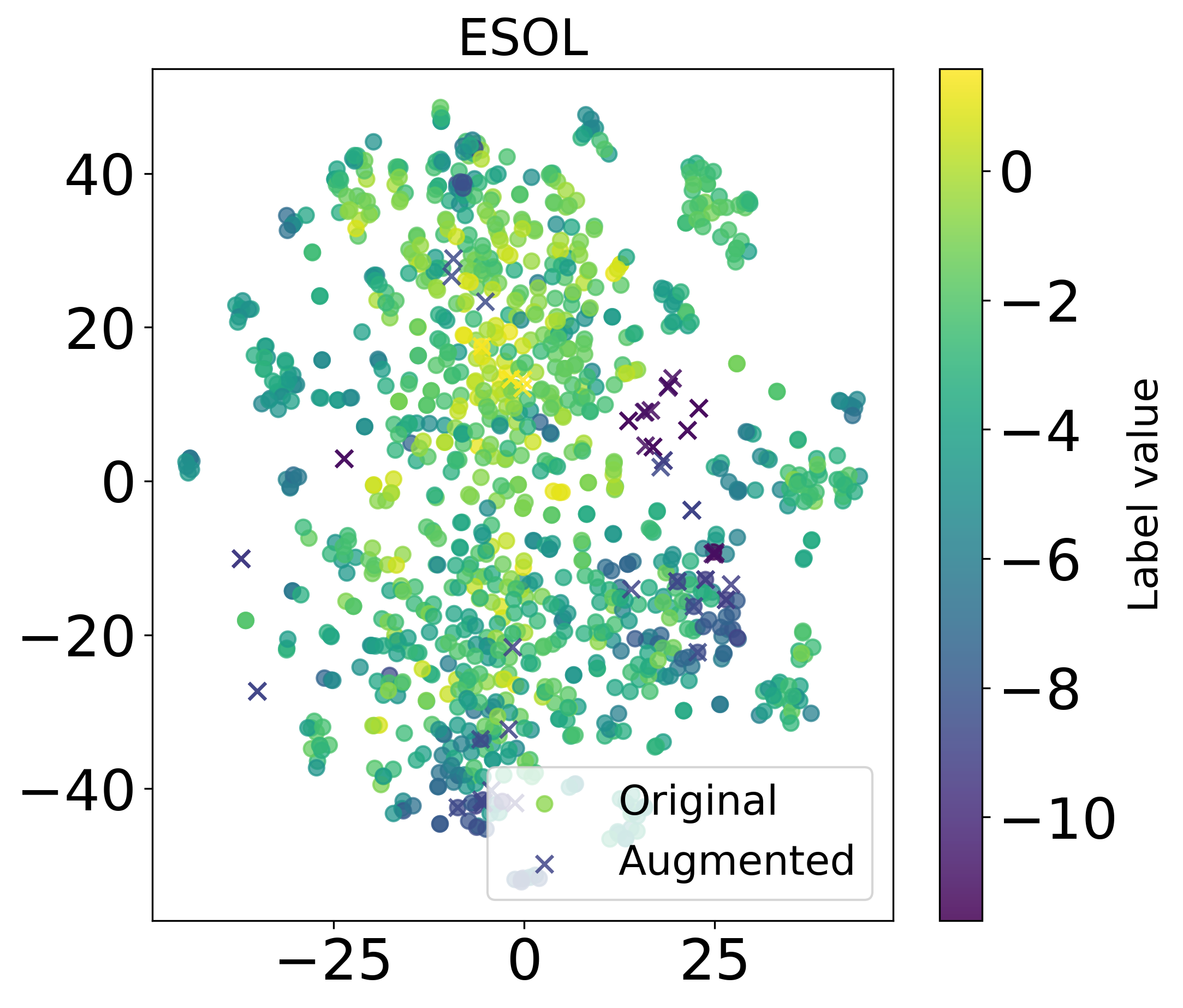}
        \end{subfigure}
        \begin{subfigure}[t]{0.49\linewidth}
            \centering
            \includegraphics[width=\linewidth]{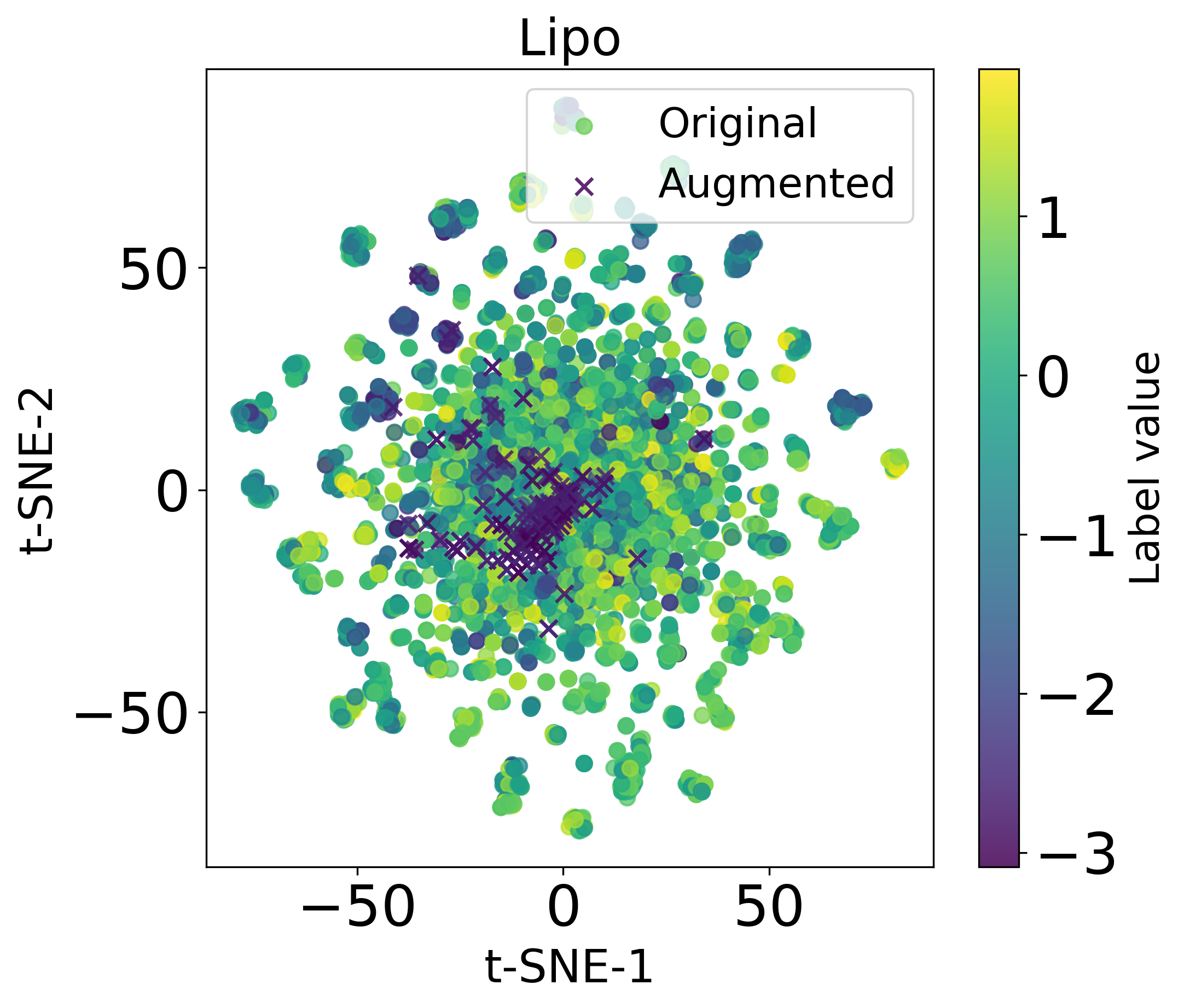}
        \end{subfigure}
        \begin{subfigure}[t]{0.49\linewidth}
            \centering
            \includegraphics[width=\linewidth]{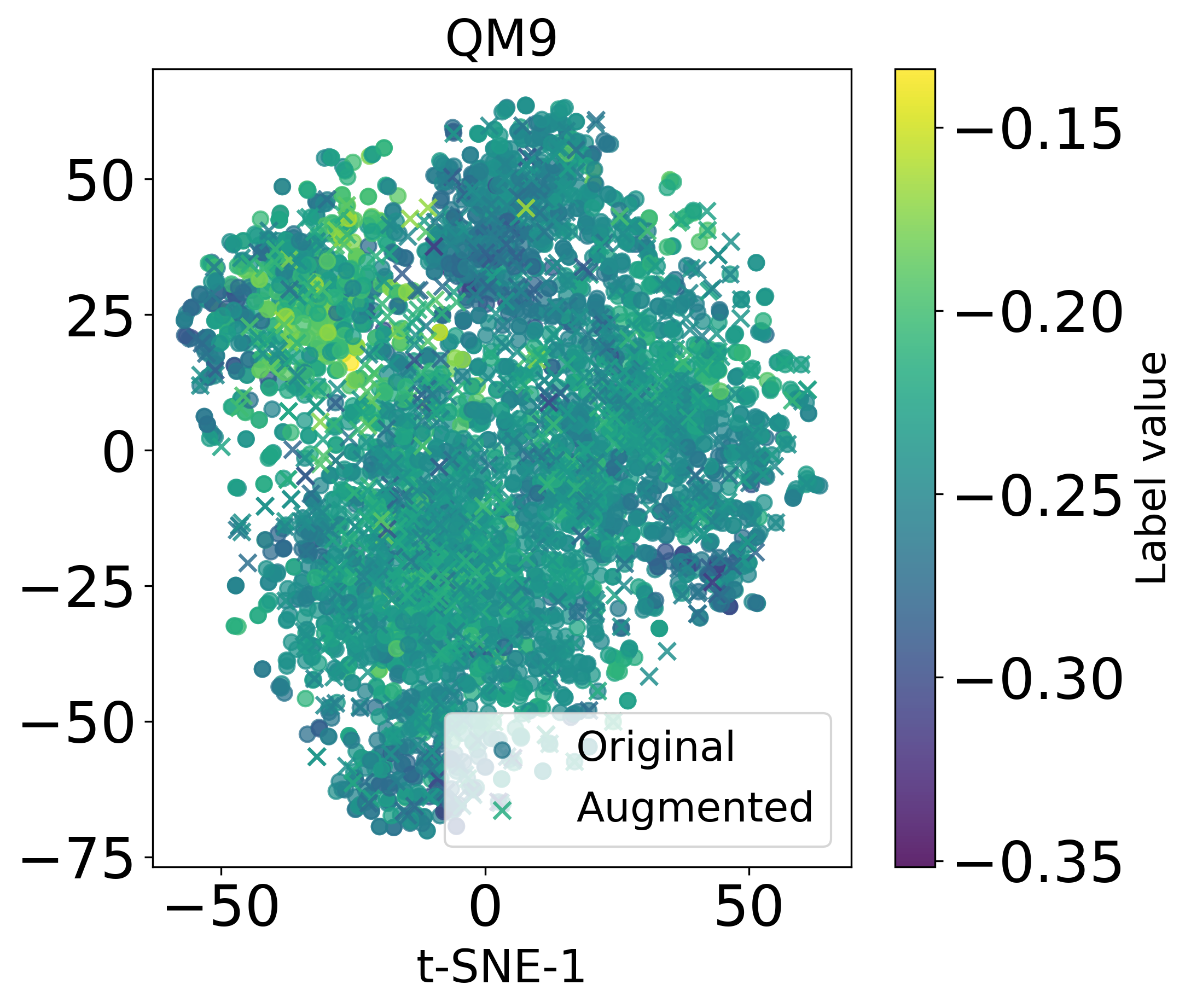}
        \end{subfigure}
        
        \caption{\textbf{t-SNE visualization of Morgan fingerprints} comparing original and augmented samples for each dataset. QM9 plot presents 2.5\% of the generated and original dataset sampled randomly for visualization purposes.}
        \label{fig:tsne_fingerprints}
\end{figure}

\begin{figure*}[ht]
    \centering
    \includegraphics[width=1.0\linewidth]{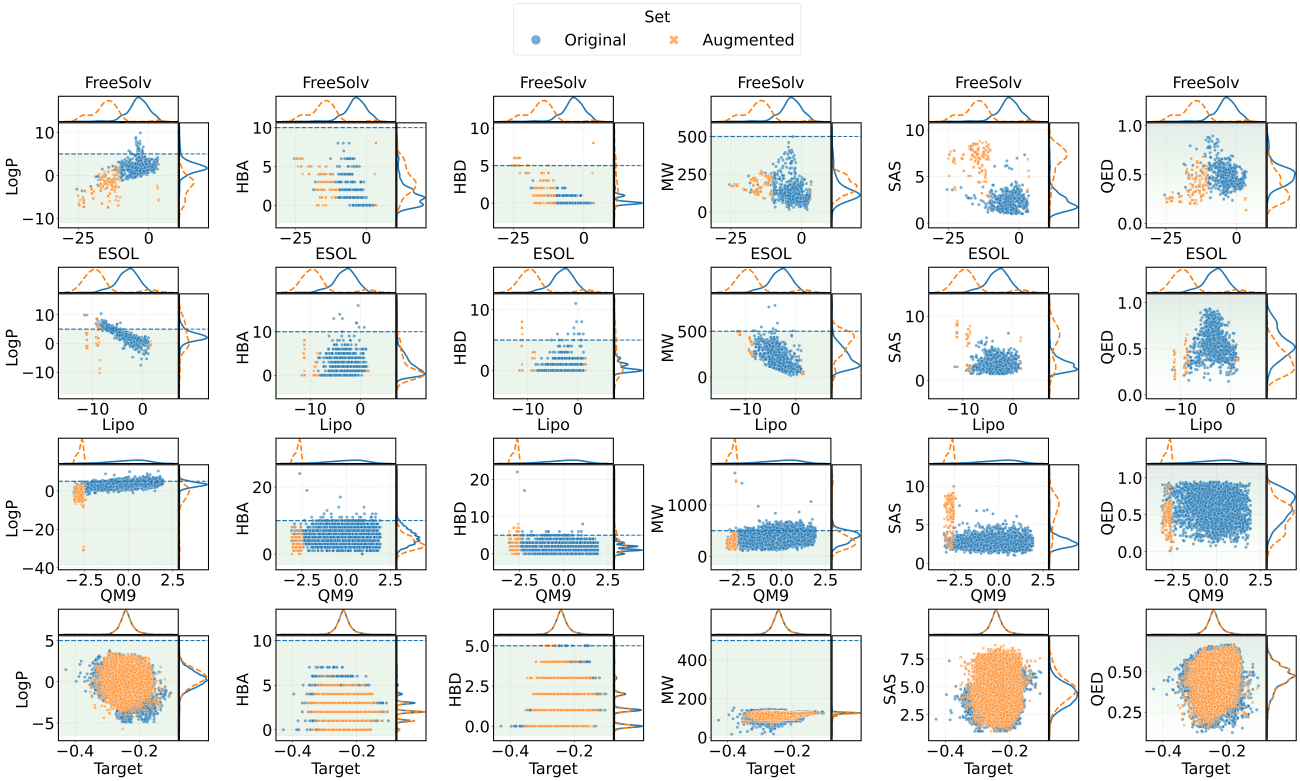}
    \caption{Joint distributions of target property values versus key physicochemical descriptors (LogP, HBA, HBD, molecular weight, synthetic accessibility score (SAS), and QED) for original (blue) and augmented (orange) molecules. Shaded areas denote Lipinski-compliant ranges ($LogP\leq5$, $MW\leq500DA$, $HBD\leq5$, $HBA\leq10$), while green regions indicate favorable properties—lower SAS (easier synthesis) and higher QED (greater drug-likeness).}
    \label{fig:fiverule}
\end{figure*}

To further understand the impact on chemical space, we visualize original and augmented molecules using t-SNE on Morgan fingerprints (Figure~\ref{fig:tsne_fingerprints}). For QM9 and ESOL, augmented samples overlap the original manifold and preserve smooth label gradients, indicating realistic local perturbations. In contrast, FreeSolv and Lipo exhibit new clusters formed by augmented samples that align with label values, suggesting exploration of property space. These results indicate that \textbf{Generated molecules populate sparse regions, improving coverage and mitigating distributional imbalance.} 


\renewcommand{\arraystretch}{0.85} 
\setlength{\tabcolsep}{3pt}        

\begin{table}[!h]
\centering
\caption{Atom, ring and bonds statistics for Original vs. Generated molecules across datasets.}\label{tab:molecules_stats}
\resizebox{1.0\linewidth}{!}{
\begin{tabular}{|ll|ccc|ccc|ccc|}
\hline
\multirow{2}{*}{\textbf{Dataset}} & \multirow{2}{*}{} & \multicolumn{3}{|c}{{Atoms}} & \multicolumn{3}{|c}{{Rings}} & \multicolumn{3}{|c|}{{Bonds}} \\ \cline{3-11} 
                                    &                                                        & $_{min}$    & $_{mean}$     & $_{max}$   & $_{min}$    & $_{mean}$    & $_{max}$    & $_{min}$    & $_{mean}$     & $_{max}$   \\ \hline
\multirow{2}{*}{FreeSolv}           & Orig             & 1      & 8.73     & 24    & 0      & 0.66    & 5      & 0      & 8.39     & 25    \\
                                    & Gen                     & 5      & 12.06    & 20    & 0      & 1.77    & 8      & 4      & 12.70    & 23    \\ \hline
\multirow{2}{*}{ESOL}               & Orig              & 1      & 13.28    & 55    & 0      & 1.39    & 8      & 0      & 13.67    & 62    \\
                                    & Gen                               & 2      & 19.83    & 28    & 0      & 2.86    & 7      & 1      & 21.69    & 31    \\ \hline
\multirow{2}{*}{Lipo}               & Orig              & 7      & 27.04    & 115   & 0      & 3.49    & 13     & 7      & 29.50    & 118   \\
                                    & Gen                          & 10     & 22.80    & 115   & 0      & 3.25    & 9      & 10     & 24.84    & 119   \\ \hline
\multirow{2}{*}{QM9}                & Orig        & 1      & 8.80     & 9     & 0      & 1.74    & 8      & 0      & 9.40     & 13    \\
                                    & Gen                      & 4      & 8.84     & 9     & 0      & 1.86    & 8      & 3      & 9.54     & 13    \\ \hline
\end{tabular}}

\end{table}

We also assess structural complexity by comparing atom, ring, and bond statistics between original and augmented molecules in Table~\ref{tab:molecules_stats}. In FreeSolv and ESOL, augmented molecules are generally larger and more cyclic, presenting higher mean atom and ring counts, along with higher bond counts, indicating increased structural complexity. Lipo and QM9, on the contrary, show similar overall distributions. Figure~\ref{fig:structural_feat} further reveals composition-level changes beyond size. FreeSolv exhibits increased ring diversity, with 6-member rings decreasing and smaller rings emerging. ESOL shows the strongest chemical shift, with aromaticity increasing and heteroatom content rising substantially, indicating more conjugated and diverse scaffolds. Lipo displays moderate redistribution toward more saturated structure, with single bonds increasing and aromaticity decreasing. 
QM9 is the only exception, exhibiting nearly unchanged distributions due to its construction - molecules are restricted to only five elements (H, C, N, O, F) - resulting in small, chemically homogeneous structures with limited ring and bond diversity. Overall, these results indicate that \textbf{our method broadens scaffold diversity while remaining within chemically reasonable bounds, reinforcing the improved dataset coverage and property–target alignment}.


Furthermore, we evaluated the generated drugs against Lipinski’s Rule of Five. These rules include drugs having five or fewer hydrogen bond donors (HBD), a molecular weigh (MW) of less than 500 Da, a partition coefficient (logP)~\cite{wildman1999prediction} of less than five, and ten or fewer hydrogen bond acceptors (HDA). As can be seen in Figure~\ref{fig:fiverule} that most of the generated molecules satisf Lipinski’s Rule of Five. The results illustrate that the joint distributions of SAS and QED, showing that the generated molecules generally fall within acceptable ranges for both properties. While some generated samples exhibit higher SAS or lower QED (suggesting increased synthetic complexity and reduced drug-likeness) the overall distribution still contains many promising candidates with low SA and high QED~\cite{ertl2009estimation,li2024drugmetric}. \textbf{This indicates that, despite variability, the generative process is capable of producing molecules that maintain desirable medicinal chemistry characteristics}.

\subsection{Ablation Study (RQ2)}

To disentangle the contributions of each design choice, we perform an ablation study on interpolation (Inter), spectral alignment (FGW), and KDE-based augmentation (Table~\ref{tab:ablation}). Removing interpolation and simply copying samples results in a consistent drop in performance across all datasets, indicating that sample diversity is essential. Excluding FGW alignment also reduces performance--except on ESOL--highlighting the importance of geometry-aware alignment for structurally diverse molecules. Likewise, removing KDE-based augmentation degrades performance in imbalanced regions, underscoring the value of density-aware augmentation for improved generalization. Overall, \textbf{the full model achieves the best or near-best performance across all datasets}. Figure~\ref{fig:paramater_search} in Appendix~\ref{app:parameters} shows the effect of the alignment weight~$\alpha$ and mixing coefficient~$\gamma$, with the best results obtained for $0.25 \le \alpha \le 0.75$ and $\gamma \in {0.1, 0.5}$. This demonstrates that \textbf{balanced alignment and coefficient jointly contribute to performance gains, confirming that each component of the pipeline plays a distinct and influential role}.

\begin{table*}[ht]

\centering
\caption{Ablation study with incremental addition of augmentation (Aug), alignment (Align), and KDE prior. Results are reported as $mean (std)_{[MeanRunTime]}$ of per-sample errors over multiple runs. Best results per dataset are highlighted in bold.} \label{tab:ablation}
\resizebox{\linewidth}{!}{
\begin{tabular}{cccc|cccc}
\toprule
Aug & Inter & Align & KDE & ESOL & FreeSolv & Lipo & QM9 \\
\midrule
$\times$ & $\times$ & $\times$ & $\times$ & 0.586(0.568) [87.84s] & 0.926(1.125) [43.03s] & 0.408(0.384) [334.27s] & 0.00369(0.005) [6980.84s] \\
$\checkmark$ & $\times$ & $\times$ & $\times$ & 0.542(0.516) [57.98s] & 0.799(1.087) [48.04s] & 0.383(0.361) [479.52s] & 0.00360(0.005) [7726.88s] \\
$\checkmark$ & $\times$ & $\times$ & $\checkmark$ & 0.542(0.516) [75.43s] & 0.799(1.087) [52.18s] & 0.383(0.361) [355.64s] & 0.00360(0.005) [7292.71s] \\
$\checkmark$ & $\checkmark$ & $\times$ & $\times$ & 0.542(0.512) [75.58s] & 0.863(1.128) [58.28s] & 0.384(0.371) [372.92s] & 0.00369(0.005) [7109.36s] \\
$\checkmark$ & $\checkmark$ & $\times$ & $\checkmark$ & \textbf{0.534(0.515) [84.42s]} & 0.869(1.113) [57.26s] & 0.378(0.362) [380.28s] & 0.00374(0.005) [7046.01s]  \\
$\checkmark$ & $\checkmark$ & $\checkmark$ & $\times$ & 0.571(0.557) [95.20s] & 0.807(1.164) [55.44s] & 0.389(0.363) [679.49s] & 0.00370(0.005) [8323.45s]  \\
$\checkmark$ & $\checkmark$ & $\checkmark$ & $\checkmark$ &\textbf{0.534(0.525) [95.90s]} & \textbf{0.769(1.023) [113.99s]} & \textbf{0.377(0.359) [691.24s]} & \textbf{0.00357(0.005) [11582.16s]} \\
\bottomrule
\end{tabular}}

\end{table*}

\subsection{Predictive Performance (RQ3)}
We compare our proposed method (\textbf{SPECTRA}) against representative state-of-the-art molecular representation learning models, including contrastive methods (GraphCL~\citep{you2020graph}, MolCLR~\citep{wang2022molecular}), language–graph hybrids (Molformer~\cite{ross2022large}, Cheb~\cite{defferrard2016convolutional}), and recent GNN-based frameworks (HiMol~\citep{zang2023hierarchical}, SGIR~\citep{liu2023semi}).  Table~\ref{tab:mae-sera} reports both the general prediction accuracy, measured by mean absolute error (MAE), and performance under imbalance, measured by squared error relevance area (SERA) ~\cite{ribeiro2020imbalanced}.  Table~\ref{tab:stats}, in the Appendix, reports pairwise statistical tests demonstrating that SPECTRA significantly outperforms most models across the datasets, with only a few cases where differences are negligible. The experimental setup is presented in Appendix~\ref{app:imple}.

\begin{table*}[ht]
\centering
\caption{Mean absolute error (MAE) and SERA with variance for each model across four datasets. Lower values indicate better performance. Bold models are the best results while underlined denote second best.}
\label{tab:mae-sera}
\resizebox{\linewidth}{!}{%
\begin{tabular}{lcccccccc}
\toprule
\textbf{Model} 
& \multicolumn{4}{c}{\textbf{MAE (mean $\pm$ var)}} 
& \multicolumn{4}{c}{\textbf{SERA (mean $\pm$ var)}} \\
\cmidrule(lr){2-5} \cmidrule(lr){6-9}
& ESOL & FreeSolv & Lipo & QM9 
& ESOL & FreeSolv & Lipo & QM9 \\
\midrule
Cheb     
& 0.59 $\pm$ 0.32 
& 0.93 $\pm$ 1.26 
& 0.41 $\pm$ 0.15 
& 0.00360 $\pm$ 0.0000
& 0.25 $\pm$ 0.01 
& 1.07 $\pm$ 0.19 
& 0.11 $\pm$ 0.00 
& \textbf{0.00002} $\pm$ \textbf{0.0000} \\

GraphCL    
& 0.78 $\pm$ 0.40 
& 1.76 $\pm$ 2.30 
& 0.73 $\pm$ 0.30 
& 0.01077 $\pm$ 0.0001
& 0.36 $\pm$ 0.00 
& 2.59 $\pm$ 0.45 
& 0.40 $\pm$ 0.00 
& 0.00011 $\pm$ 0.0000 \\

HiMol      
& \underline{0.51} $\pm$ \underline{0.22} 
& 0.97 $\pm$ 1.46 
& 0.41 $\pm$ 0.14 
& 0.06296 $\pm$ 0.0018
& \underline{0.17} $\pm$ \underline{0.00} 
& 1.34 $\pm$ 0.74 
& 0.10 $\pm$ 0.00 
& 0.00183 $\pm$ 0.0000 \\

MolCLR     
& 0.73 $\pm$ 0.40 
& 1.12 $\pm$ 1.31 
& 0.43 $\pm$ 0.14 
& 0.00364 $\pm$ \underline{0.0000}
& 0.32 $\pm$ 0.00 
& 1.36 $\pm$ 0.18 
& 0.11 $\pm$ 0.00 
&\textbf{0.00002} $\pm$ \textbf{0.0000} \\

Molformer  
& 1.66 $\pm$ 1.68 
& 2.84 $\pm$ 6.87 
& 0.81 $\pm$ 0.38 
& 0.01628 $\pm$ 0.0002
& 2.77 $\pm$ 0.01 
& 10.61 $\pm$ 12.16 
& 0.54 $\pm$ 0.00 
& 0.00038 $\pm$ 0.0000 \\

SGIR       
& \textbf{0.46} $\pm$ \textbf{0.19} 
& \textbf{0.68} $\pm$ \textbf{0.85} 
& \textbf{0.37} $\pm$ \textbf{0.13} 
& \textbf{0.00341} $\pm$ \textbf{0.0000}
& \textbf{0.13} $\pm$ \textbf{0.00} 
& \textbf{0.69} $\pm$ \textbf{0.05} 
& \textbf{0.09} $\pm$ \textbf{0.00} 
& \textbf{0.00002} $\pm$ \textbf{0.0000} \\ \hline

SPECTRA    
& 0.53 $\pm$ 0.28 
& \underline{0.77} $\pm$ \underline{1.05} 
& \underline{0.38} $\pm$ \underline{0.13} 
& \underline{0.00357} $\pm$ \underline{0.0000}
& 0.20 $\pm$ 0.00 
& \underline{0.95} $\pm$ \underline{0.29} 
& \textbf{0.09} $\pm$ \textbf{0.00} 
& \textbf{0.00002} $\pm$ \textbf{0.0000} \\
\bottomrule
\end{tabular}%
}

\end{table*}

\begin{figure}[!h]

    \centering
    \includegraphics[width=\linewidth]{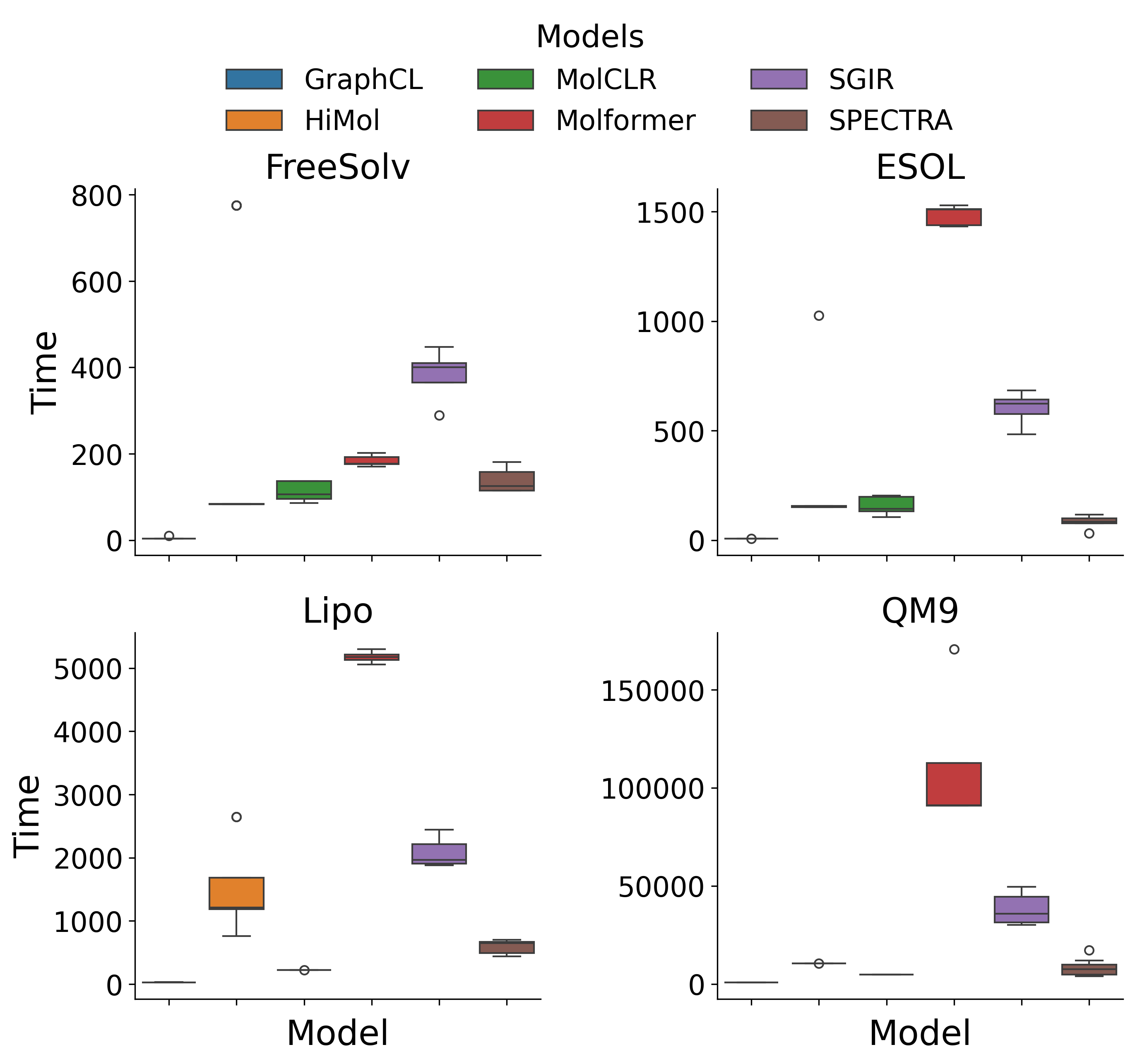}
    \caption{%
    Boxplots compare the inference/training runtime of the models on datasets. Each subplot corresponds to one dataset, and box distributions summarize the variability of runtime on train across 5 folds cross-validation.
    }
    \label{fig:time}
\end{figure}
\textbf{SPECTRA achieves consistently strong performance across all datasets}. On ESOL, Lipo, and QM9, it remains competitive with the best-performing models, while on FreeSolv it surpasses most baselines, on both metrics, obtaining the second-best overall performance. Although SGIR attains the lowest MAE and SERA on average, these gains come at the cost of a substantially heavier training pipeline, as it relies on pseudo-labeling and latent-space augmentation. In contrast, \textbf{SPECTRA provides a more efficient and stable alternative, maintaining competitive accuracy while achieving strong performance in underrepresented regions}, as reflected by its consistently low SERA values. These results indicate that our spectral alignment generation strategy improves prediction in imbalanced regimes without sacrificing scalability or practical usability. 

\vspace{-2.6mm}\subsection{Time Efficiency Analysis (RQ4)}~\label{sec:time}Computational efficiency is also crucial for real-world deployment. Figure~\ref{fig:time} shows the runtime distribution of all models across the three datasets. \textbf{SPECTRA is time-efficient and comparable to the fastest methods evaluated, and it runs substantially faster than SGIR (2x-7x faster)}.

\begin{figure}[!h]

    \centering
    \includegraphics[width=\linewidth]{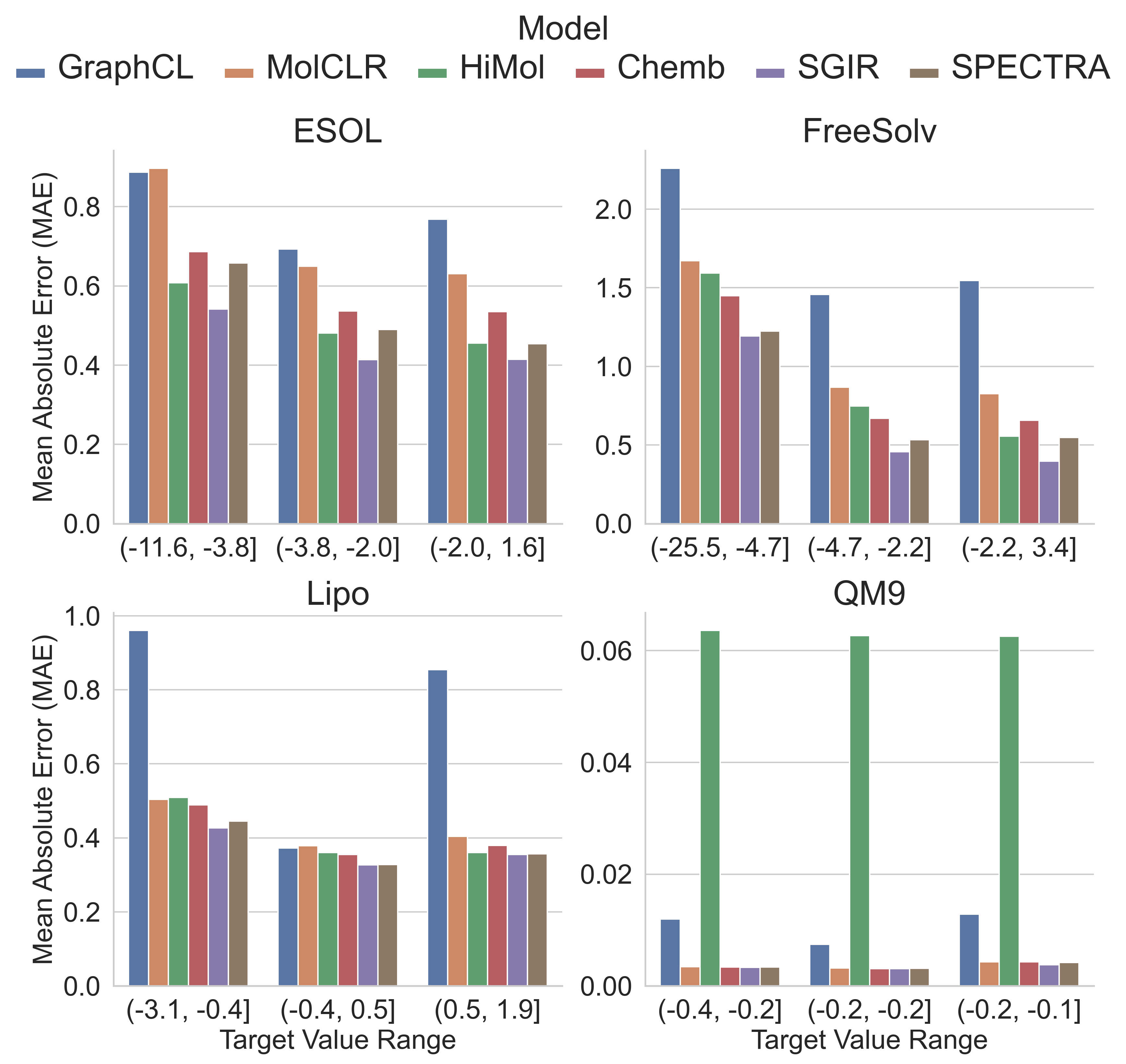}
    \caption{\textbf{Mean Absolute Error (MAE) distribution} across target value ranges for each dataset. 
   Colors correspond to different models as indicated in the legend.}
   \label{fig:errors_dist}
\end{figure}

\subsection{Error Distribution Across Domain (RQ5)}
Figure~\ref{fig:errors_dist} further dissects MAE by target value ranges. We observe that baseline models often suffer from considerably higher errors in the low-density regions, consistent with the imbalance in the training data.  In contrast, SPECTRA demonstrates markedly lower errors in these sparse regions, highlighting its strength in addressing imbalance without degrading performance in high-density regions.

\section{Conclusion}
Experiments across benchmark datasets demonstrate that our method improves predictive accuracy in rare but critical property regimes, preserves structure–property correlations, and achieves a favorable balance between accuracy and computational efficiency. These results establish spectral-based generation as a promising and interpretable strategy for addressing class imbalance in molecular property prediction and related graph-structured scientific domains.

Several directions remain open for future work. First, we plan to extend the framework to multi-property prediction tasks and incorporate additional modalities such as 3D geometric features. Second, we aim to reduce the computational cost of Fused Gromov-Wasserstein alignment and spectral decomposition, either through more efficient algorithmic variants or hybrid strategies that combine spectral augmentation with model-based pseudo-labeling to better capture non-linear structure–label relationships. Finally, we intend to evaluate the approach on molecular datasets beyond MoleculeNet to assess generalization across domains. 

\section{Limitations and Ethical Considerations}

For the small, drug-like molecules typical of MoleculeNet benchmarks, Fused Gromov–Wasserstein alignment remains computationally tractable. However, its cost may become prohibitive for larger and more complex graphs, such as polymers or protein–ligand systems. Addressing this limitation will require the integration of more efficient or approximate Optimal Transport solvers. In addition, our current framework is restricted to non-radical molecules, and extending it to handle radicals represents an important direction for future work

This work is intended to improve predictive modeling for scientific discovery. As with any generative augmentation method, the synthetic molecules produced should be treated as computational hypotheses subject to experimental validation, not as candidates for direct deployment. Practitioners should also be aware that imbalanced training sets often reflect real-world scarcity of certain chemical classes, and augmentation cannot substitute for domain knowledge about underrepresented regions of the chemical space.

\begin{acks}
The project was partially supported by the National Science Foundation under the NSF Center for Computer Assisted Synthesis (C-CAS; grant no. CHE-2202693).
\end{acks}

\section*{GenAI Disclosure}

We have used Generative AI in this work for English correction and code checking. 
\bibliographystyle{ACM-Reference-Format}
\bibliography{bib_tex}

\appendix

\section{Dataset Details}\label{app:dataset}

Our experimental evaluation uses molecular regression tasks from MoleculeNet~\citep{wu2018moleculenet}, specifically ESOL, FreeSolv, Lipophilicity (Lipo) and QM9. A brief summary of these datasets is provided in Table~\ref{tab:dataset}. The QM9 dataset provides 12 quantum-chemical properties, from which we selected the HOMO energy as the target for evaluation.

\begin{table}[!h]
\centering
\caption{Summary of Molecular Property Datasets}
\label{tab:dataset}
\resizebox{1.0\linewidth}{!}{
\begin{tabular}{lcc}
\toprule
\textbf{Dataset} & \textbf{\# of Compounds} & \textbf{Description} \\
\midrule
ESOL           & 1{,}128 & Water solubility (log solubility in mol/L) \\
FreeSolv       & 642     & Hydration free energy in water \\
Lipophilicity  & 4{,}200 & Octanol/water distribution coefficient (logD at pH 7.4) \\
QM9           & 133{,}886 & Geometric, energetic, electronic and thermodynamic\\
    & &       properties of DFT-modelled small molecules. \\
\bottomrule
\end{tabular}
}
\end{table}


\section{Parameter Evaluation}\label{app:parameters}
\begin{figure}[]
    \centering
    \includegraphics[width=0.65\linewidth]{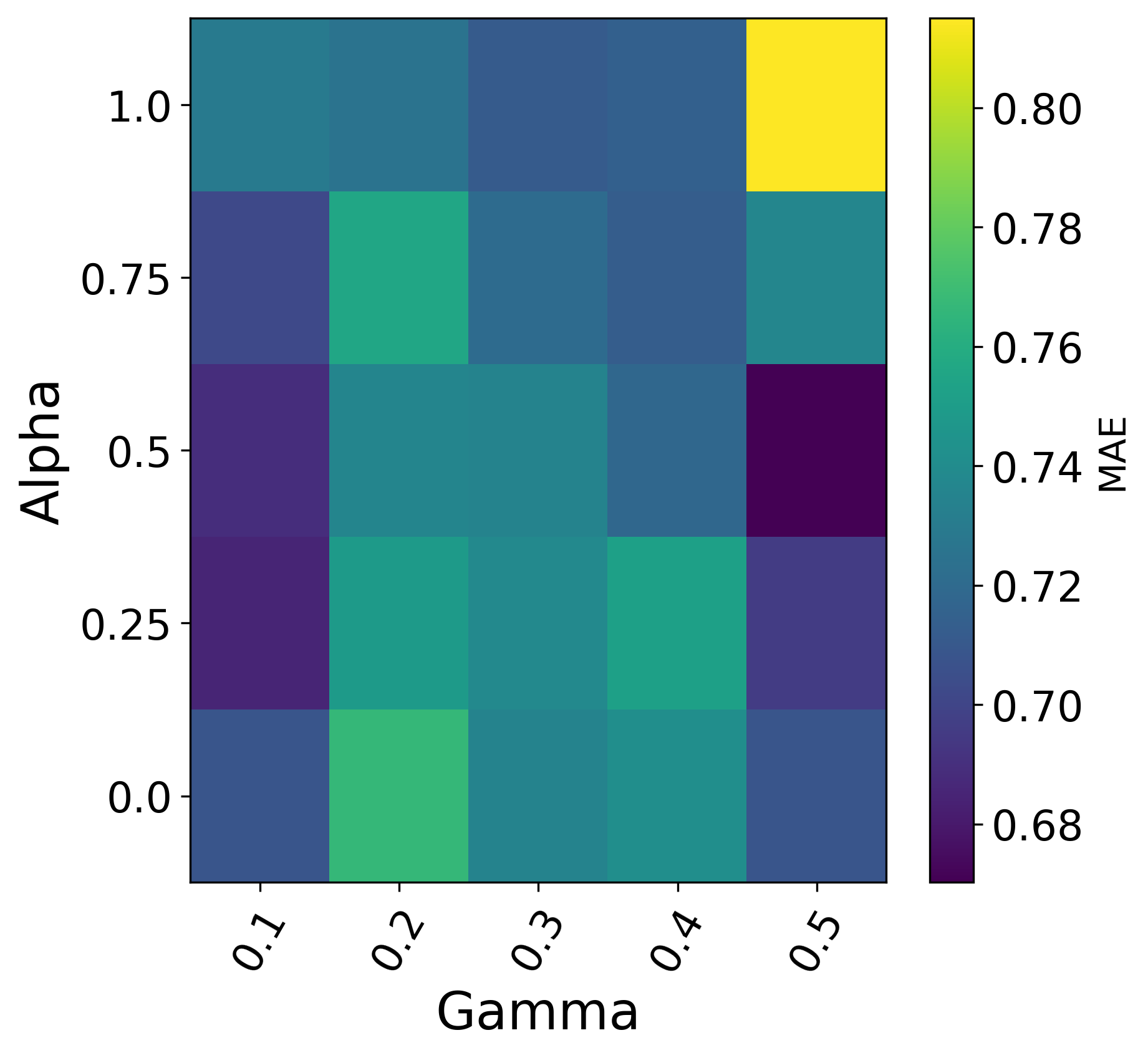}
    \caption{MAE surfaces for the $\alpha$ and $\gamma$ hyperparameter in the FreeSolv dataset. Each heatmap shows the MAE computed on the validation set, with $\alpha$ balancing structural features for alignment and $\gamma$ as the mixing coefficient.}
    \label{fig:paramater_search}
\end{figure}

Figure~\ref{fig:paramater_search} presents the impact of different value combinations of parameters in the validation sets, exploring the joint effects of the alignment parameter~$\alpha$ (which controls the balance of structural features) and the mixing coefficient~$\gamma$. The heatmap summarizes the corresponding mean absolute error (MAE) across combinations of $(\alpha,\gamma)$, allowing visual inspection of how these parameters interact and influence predictive performance. We choose lower values for $\gamma$ as we want to be closer to the sample chosen for augmentation. 

\section{Statistical Improvement}\label{app:stats}
Table \ref{tab:stats} provides a detailed statistical comparison between SPECTRA and all baseline models across the ESOL, FreeSolv, Lipo, and QM9 datasets. For each pairwise comparison, we compute the raw absolute-error improvement, defined as the per-sample difference in absolute prediction errors between SPECTRA and the competing model. Positive values indicate that SPECTRA produces lower predictive error. To assess whether these differences are statistically meaningful, we apply the paired t-test. The “Best Model” column identifies which method achieves the lowest absolute errors on average, while the “Significant” column reports whether the observed improvement is statistically significant.

\begin{table}[]

\centering
\caption{Statistical comparison between SPECTRA and baseline models across all datasets. The table reports raw absolute-error improvements (positive values indicate lower error for SPECTRA), T-test p-values, the best-performing model for each dataset comparison, and whether the difference is statistically significant at $\alpha = 0.05$.}\label{tab:stats}
\resizebox{1.0\linewidth}{!}{%
\begin{tabular}{lllccrr}
\toprule
 & Dataset & Model & t-test & \makecell{Mean \\ Improvement} & Best Model & Significant \\
\midrule
0 & esol & Cheb & 0.000031 & 0.051921 & SPECTRA & Yes \\
1 & esol & GraphCL & 0.000000 & 0.248316 & SPECTRA & Yes \\
2 & esol & HiMol & 0.212130 & -0.019109 & HiMol & No \\
3 & esol & MolCLR & 0.000000 & 0.191656 & SPECTRA & Yes \\
4 & esol & Molformer & 0.000000 & 1.123755 & SPECTRA & Yes \\
5 & esol & SGIR & 0.000000 & -0.077233 & SGIR & Yes \\
6 & freesolv & Cheb & 0.000000 & 0.157728 & SPECTRA & Yes \\
7 & freesolv & GraphCL & 0.000000 & 0.987215 & SPECTRA & Yes \\
8 & freesolv & HiMol & 0.000002 & 0.197025 & SPECTRA & Yes \\
9 & freesolv & MolCLR & 0.000000 & 0.353207 & SPECTRA & Yes \\
10 & freesolv & Molformer & 0.000000 & 2.073809 & SPECTRA & Yes \\
11 & freesolv & SGIR & 0.034940 & -0.085648 & SGIR & Yes \\
12 & qm9 & Cheb & 0.000000 & 0.000037 & SPECTRA & Yes \\
13 & qm9 & GraphCL & 0.000000 & 0.007201 & SPECTRA & Yes \\
14 & qm9 & HiMol & 0.000000 & 0.059398 & SPECTRA & Yes \\
15 & qm9 & MolCLR & 0.000009 & 0.000077 & SPECTRA & Yes \\
16 & qm9 & Molformer & 0.000000 & 0.012717 & SPECTRA & Yes \\
17 & qm9 & SGIR & 0.000000 & -0.000161 & SGIR & Yes \\
18 & lipo & Cheb & 0.000000 & 0.031565 & SPECTRA & Yes \\
19 & lipo & GraphCL & 0.000000 & 0.352521 & SPECTRA & Yes \\
20 & lipo & HiMol & 0.000000 & 0.033232 & SPECTRA & Yes \\
21 & lipo & MolCLR & 0.000000 & 0.052444 & SPECTRA & Yes \\
22 & lipo & Molformer & 0.000000 & 0.430608 & SPECTRA & Yes \\
23 & lipo & SGIR & 0.153037 & -0.006892 & SGIR & No \\
\bottomrule
\end{tabular}}

\end{table}

Overall, the results demonstrate that SPECTRA consistently outperforms most baselines, with particularly strong margins on FreeSolv, showing both large error reductions and highly significant $p$-values. A few cases, such as HiMol and SGIR on specific datasets, show either negligible differences or slight advantages for the baseline, and these are reflected accordingly in the statistical tests.

\begin{table*}[ht]
\centering
\caption{Mean absolute error (MAE) and SERA with variance for each model across four datasets. Lower values indicate better performance. Bold denotes the best overall results; underlined denotes the best within each model pair.}
\label{tab:mae-sera-downstream}
\resizebox{0.93\linewidth}{!}{%
\begin{tabular}{lcccccccc}
\toprule
\textbf{Model} 
& \multicolumn{4}{c}{\textbf{MAE (mean $\pm$ var)}} 
& \multicolumn{4}{c}{\textbf{SERA (mean $\pm$ var)}} \\
\cmidrule(lr){2-5} \cmidrule(lr){6-9}
& ESOL & FreeSolv & Lipo & QM9 
& ESOL & FreeSolv & Lipo & QM9 \\
\midrule
ChemBERTa
& \underline{0.79} $\pm$ \underline{0.53} & \underline{1.19} $\pm$ \underline{2.54} & \underline{0.49} $\pm$ \underline{0.20} & \underline{0.00492} $\pm$ \underline{0.0000}
& \underline{0.47} $\pm$ \underline{0.00} & \underline{2.72} $\pm$ \underline{2.34} & \underline{0.17} $\pm$ \underline{0.00} & \underline{0.00003} $\pm$ \underline{0.0000} \\
ChemBERTa + SPECTRA
& 0.80 $\pm$ 0.50 & 1.24 $\pm$ 2.58 & 0.49 $\pm$ 0.20 & 0.00512 $\pm$ 0.0000
& 0.48 $\pm$ 0.00 & 2.80 $\pm$ 2.23 & 0.17 $\pm$ 0.00 & \underline{0.00003} $\pm$ \underline{0.0000} \\ \hline
GCN
& 0.59 $\pm$ 0.31 & \underline{0.81} $\pm$ \underline{1.05} & 0.42 $\pm$ 0.15 & \underline{0.00443} $\pm$ \underline{0.0000}
& 0.24 $\pm$ 0.00 & \textbf{0.90} $\pm$ \underline{0.20} & 0.12 $\pm$ 0.00 & \textbf{0.00002}$\pm$ \textbf{0.0000} \\
GCN + SPECTRA
& \underline{0.55} $\pm$ \underline{0.28} & 0.83 $\pm$ 1.30 & \underline{0.41} $\pm$ \underline{0.15} & 0.00449 $\pm$ 0.0000
& \underline{0.23} $\pm$ \underline{0.00} & 1.10 $\pm$ 0.29 & \underline{0.11} $\pm$ \underline{0.00} & \textbf{0.00002}$\pm$ \textbf{0.0000} \\ \hline
Cheb
& 0.59 $\pm$ 0.32 & 0.93 $\pm$ 1.26 & 0.41 $\pm$ 0.15 & 0.00360 $\pm$ 0.0000
& 0.25 $\pm$ 0.01 & 1.07 $\pm$ 0.19 & 0.11 $\pm$ 0.00 & \textbf{0.00002} $\pm$ \textbf{0.0000} \\
\textbf{Cheb + SPECTRA (Ours)}
& \textbf{0.53} $\pm$ \textbf{0.28} & \textbf{0.77} $\pm$ \textbf{1.05} & \textbf{0.38} $\pm$ \textbf{0.13} & \textbf{0.00357} $\pm$ \textbf{0.0000}
& \textbf{0.20} $\pm$ \textbf{0.00} & \underline{0.95} $\pm$ \underline{0.29} & \textbf{0.09} $\pm$ \textbf{0.00} & \textbf{0.00002} $\pm$ \textbf{0.0000} \\
\bottomrule
\end{tabular}%
}
\end{table*}
\begin{figure*}
    \centering
    \includegraphics[width=0.94\linewidth]{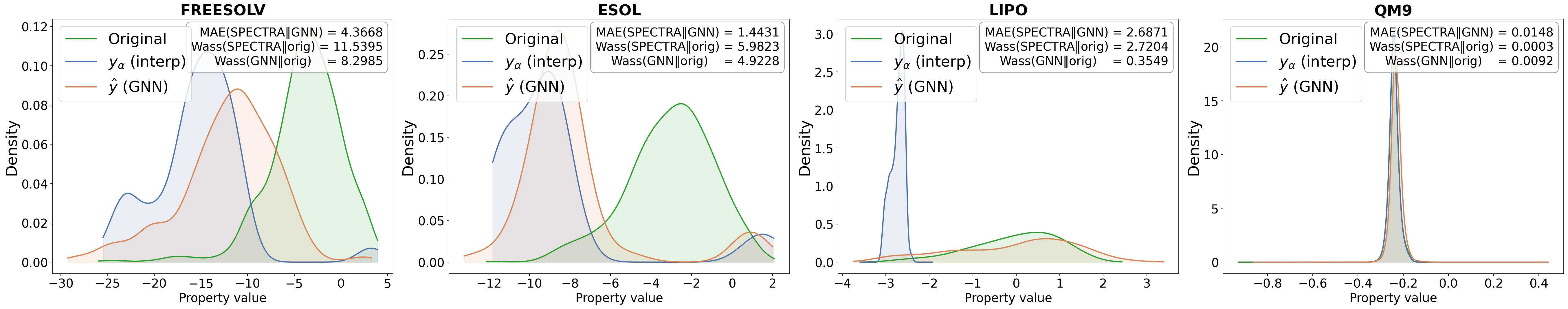}
    \caption{Property value distributions for original training molecules, interpolated labels, and GNN-predicted labels alongside MAE values between interpolated values and GNN and Wasserstein distance from original training molecules.}
    \label{fig:label_validation}
\end{figure*}
\section{Experimental Setup}\label{app:imple}

All experiments were conducted on a Linux server equipped with two 12-core \texttt{Intel(R) Haswell} processors, 256 GB of RAM, and four \texttt{NVIDIA A100} GPUs, each with 80 GB of memory.  
Our method is implemented in \texttt{Python~3.8.19} using \texttt{PyTorch~2.1.2}.  We used Chebyshev GCN (cheb)~\citep{defferrard2016convolutional}. We perform a manual hyperparameter search over the following ranges:
\begin{itemize}
    \item \textbf{Hidden dimension:} $\{128,\,256,\,512\}$
    \item \textbf{Number of spectral block layers:} $\{3,\,4,\,5\}$
    \item \textbf{Dropout:} $\{0.0,\,0.1,\,0.3\}$
    \item \textbf{Learning rate:} $\{10^{-3},\,2\times10^{-3},\,5\times10^{-4}\}$
    \item \textbf{Chebyshev filter order ($k$):} $\{2,\,3,\,5\}$
    \item \textbf{Epochs:} $\{500\}$
    \item \textbf{Batch size:} $\{32,\,64\}$
    \item \textbf{Gamma:} $\{0.1,\,0.2,\,0.3,\,0.4,\,0.5\}$
    \item \textbf{Alpha:} $\{0.1,\,0.25,\,0.5,\,0.75,\,1\}$
    \item \textbf{Augmentation Percentage(\%):} $\{0.1,0.15,0.2,0.25\}$
\end{itemize}

The optimal configuration uses a hidden dimension of 128, four layers, a dropout rate of 0.1, a learning rate of \(2\times10^{-3}\), and hyperparameters \(k = 2\), \(\alpha = 0.5\), and \(\gamma = 0.5\). The selected augmentation rates are 0.25 for FreeSolv and 0.10 for ESOL, Lipo and QM9.

\section{SPECTRA Across Prediction Models}

To investigate how SPECTRA-generated molecular graphs improve the performance of different prediction models, we conduct additional experiments combining SPECTRA-based augmentation with three downstream predictors: Cheb~\cite{defferrard2016convolutional}, GCN~\citep{zhang2019graph}, and ChemBERTa~\citep{chithrananda2020chemberta}. As shown in Table~\ref{tab:mae-sera-downstream}, the effect of SPECTRA augmentation varies across models. For GCN, SPECTRA improves performance on ESOL and Lipo. For ChemBERTa, however, the augmentation does not yield consistent gains, with slight degradation observed across several datasets --- suggesting that transformer-based models pretrained on large corpora may be less sensitive to graph-level augmentation. ChebNet, on the other hand, benefits most consistently and substantially from SPECTRA augmentation, achieving the lowest MAE and SERA values across nearly all datasets. These results demonstrate that SPECTRA is broadly applicable as an augmentation strategy, while also revealing that the choice of downstream predictor plays an important role in the magnitude of improvement. The consistent superior gains observed with ChebNet across all evaluated datasets justify our design choice.

\section{VAE Baseline}
\label{app:vae_motivation}
 
To understand the effectiveness of our generative method, we construct a VAE baseline~\citep{marchan2025vae} designed as a controlled ablation. The baseline mirrors our augmentation pipeline in every respect except the generation step: it uses the same augmentation budget, the same KDE inverse-density weighting to focus generation on underrepresented label regions, and the same five-fold cross-validation splits. In the VAE baseline, each training molecule is first encoded into a Gaussian posterior in a SELFIES-based latent space, and new samples are generated by perturbing its posterior mean with Gaussian noise before being decoded back into valid molecular structures. 
 
\section{Linear Interpolation Assumption Analysis}\label{app:label_validation}

Our augmentation method assigns property labels to generated molecules via linear interpolation. To empirically validate this assumption, we train the same \textsc{Cheb} model (Section~\ref{sec:model}) on each training fold and use it to predict properties for all augmented molecules, comparing these predictions with the interpolated labels. Figure~\ref{fig:label_validation} reports the mean absolute error (MAE) between $\hat{y}{\text{GNN}}$ and $y{\alpha}$, along with the Wasserstein distance between each label distribution and the original training distribution. We observe higher MAE values for FreeSolv and Lipo, indicating larger discrepancies between interpolated and model-predicted labels, whereas ESOL and QM9 exhibit lower MAE. In terms of distributional alignment, the GNN predictions tend to more closely match the training distribution than SPECTRA (except in QM9), particularly for Lipo. However, despite the higher MAE in FreeSolv, both SPECTRA and the GNN demonstrate alignment in underrepresented regions, a trend that is also evident in ESOL. In Lipo, our method continues to provide stronger coverage of these underrepresented regions, while the GNN follows the original training distribution.

\section{Valid Molecule Generation}

Our method adopts an exploratory strategy in which candidate molecules are generated by interpolating between structurally similar pairs. Given that linearity is a strong assumption in molecular space, we do not expect successful interpolation for all pairs, and some degree of search is inherently required. Despite this, the method consistently identifies valid molecules across datasets, even within the vast and highly discrete chemical space. The proportion of successful interpolations varies by dataset, yielding approximately 3.2\% for FreeSolv, 15.6\% for ESOL, 2.8\% for Lipophilicity, and 27.7\% for QM9. These differences reflect the influence of dataset-specific structural properties and distributional characteristics on interpolation feasibility.
\end{document}